\pdfoutput=1

\documentclass[11pt]{article}

\usepackage{acl}

\usepackage{times,graphicx}
\usepackage{latexsym,booktabs}

\usepackage{euflag}

\usepackage[T1]{fontenc}

\usepackage[utf8]{inputenc}

\usepackage{microtype}

\usepackage{booktabs}
\usepackage{multirow}

\title{Ancestor-to-Creole Transfer is Not a Walk in the Park}

\author{Heather Lent ~ Emanuele Bugliarello ~ Anders Søgaard \\
  University of Copenhagen, Denmark \\
  \texttt{\{hcl, emanuele, soegaard\}@di.ku.dk} 
}

\begin{document}
\maketitle
\begin{abstract}
We aim to learn language models for Creole languages for which large volumes of data are not readily available, and therefore explore the potential transfer from ancestor languages (the `Ancestry Transfer Hypothesis'). We find that standard transfer methods do not facilitate ancestry transfer. 
Surprisingly, different from other non-Creole languages, a very distinct two-phase pattern emerges for Creoles: As our training losses plateau, and language models begin to overfit on their source languages, perplexity on the Creoles {\em drop}.
We explore if this {\em compression} phase can lead to practically useful language models (the `Ancestry Bottleneck Hypothesis'), but also falsify this. 
Moreover, we show that Creoles even exhibit this two-phase pattern even when training on random, unrelated languages. Thus Creoles seem to be typological outliers and we speculate whether there is a link between the two observations.

\end{abstract}

\section{Introduction}\label{sec:intro}
Creole languages refer to vernacular languages, many of which developed in colonial plantation settlements in the 17th and 18th centuries. Creoles most often emerged as a result of contact between social groups that spoke mutually unintelligible languages, i.e., from the interactions of speakers of nonstandard varieties of European languages and speakers of non-European languages~\cite{lent-etal-2021-language}. 
Some argue these languages have an exceptional status among the world's languages \cite{mcwhorter1998}, while others counter that Creoles are not unique, and evolve in the typical manner as other languages \cite{Aboh2016ANT}.
In this paper, we will present experiments in evaluating language models trained on non-Creole languages for Creoles, as well as in various control settings. We first explore the following hypothesis: 

\begin{itemize}
    \item[{\bf R1:}] Language models trained on ancestor languages should transfer well to Creole languages.
\end{itemize}

\noindent We call {\bf R1} the `Ancestry Transfer Hypothesis.' Our experiments, however, suggest that {\bf R1} is {\em not} easily validated. We note, though, that ancestor-to-Creole training exhibits divergent behavior when training {\em for long}, leading to the following hypothesis: 

\begin{itemize}
    \item[{\bf R2:}] Language models trained on ancestor languages can, after a compression phase, transfer well to Creole languages.
\end{itemize}

\noindent We call {\bf R2} the `Ancestry Bottleneck Hypothesis.' While compression benefits transfer, performance never seems to reach useful levels. Furthermore, similar effects are observed with Creoles when training on non-ancestor languages. Our findings here are not relevant to applied NLP, but they shed light on cross-lingual training dynamics \cite{singh-etal-2019-bert,deshpande2021bert}, and we believe they have potential implications for the linguistic study of Creoles \cite{DeGraff2005LinguistsMD}, as well as for information bottleneck theory \cite{Tishby99theinformation}. 

\begin{figure}
    \centering
    \includegraphics[width=2.8in]{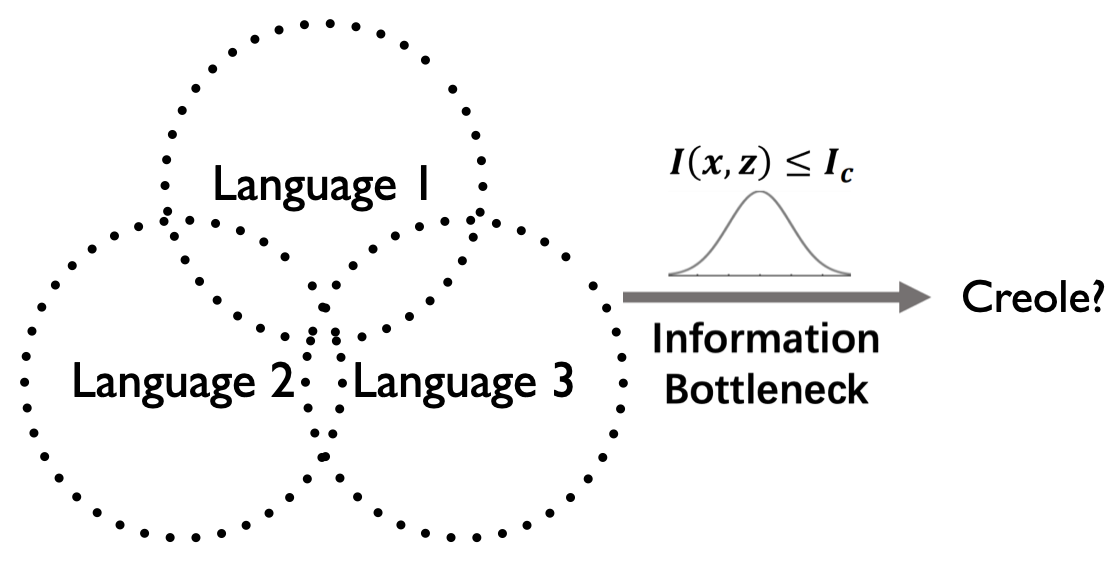}
    \caption{Does the Information Bottleneck principle capture some of the dynamics of Creole formation?}
    \label{fig:front}
\end{figure}

\begin{table*}[t]
    \centering
    \small 
    \begin{tabular}{lll}
    \toprule
    {\bf Creole}&{\bf Ancestors}&{\bf Random Controls}\\
    \midrule 
         Nigerian Pidgin &  English, Hausa, Yoruba, Portuguese & Afrikaans, Cherokee, Hungarian, Quechua\\
         Jamaican Patois & English, Hausa, Spanish, Igbo & Afrikaans, Cherokee, Hungarian, Quechua\\
         Saint Lucian Creole & French, Hausa, Yoruba, Igbo & Afrikaans, Cherokee, Hungarian, Quechua \\
         Haitian Creole & French, Fon, Spanish, Igbo & Afrikaans, Cherokee, Hungarian, Quechua \\
         \toprule
        {\bf Non-Creole}&{\bf Relatives}&{\bf Random Controls}\\
    \midrule 
        Spanish & French, Italian, Portuguese, Romanian & Afrikaans, Cherokee, Hungarian, Quechua \\
        Danish & Norweigan, Icelandic, Swedish, German & Afrikaans, Cherokee, Hungarian, Quechua \\
    \bottomrule
    \end{tabular}
    \caption{Transfer setups in our study. We aim to learn target Creoles and Non-Creoles by training on \textbf{1)} their Ancestors or Relatives, respectively; and \textbf{2)} languages unrelated to the target ones as a control (Random Controls).}
    \label{tab:lang}
    \vspace{-2mm}
\end{table*}

\paragraph{Our contributions} We conduct a large set of experiments on cross-lingual zero-shot applications of language models to Creoles, primarily to test whether ancestor languages provide useful training data for Creoles (the `Ancestry Transfer Hypothesis;' {\bf R1}). 
Our results are a mix of negative and positive results: 
{\bf First Negative Result:} Ordinary transfer methods do not enable ancestor-to-Creole transfer. 
{\bf First Positive Result:} Regardless of the source languages, when training for long periods of time, a compression phase takes places for Creoles: as the models overfit their training data, perplexity on Creoles begin to decrease. This pattern is unique to Creoles as it does not emerge for target non-Creole languages. 
{\bf Second Negative Result:} The compression phase does not lead to better representations for downstream tasks in the target Creoles.

\section{Background}
\paragraph{Cross-lingual training dynamics} Several multilingual language models have been presented and evaluated in recent years. Since \citet{singh-etal-2019-bert} showed that mBERT \cite{devlin-etal-2019-bert}  
generalizes well across related languages, but compartmentalizes language families, several researchers have explored the training dynamics of training multilingual language models across related or distant language sets \cite{lauscher-etal-2020-zero,keung-etal-2020-dont,deshpande2021bert}. Unlike most previous work on cross-lingual training, we focus on evaluation on unseen (Creole) languages. This set-up is also explored in previous work focusing on generalization to unseen scripts \cite{muller-etal-2021-unseen,pfeiffer-etal-2021-unks}. \citet{muller-etal-2021-unseen} argue that generalization to unseen languages is possible for seen scripts, but hard or impossible for unseen scripts, but this paper identifies a third category of unseen languages with seen scripts, which exhibit non-traditional learning curves in the zero-shot pre-training regime. 

\paragraph{Linguistic theories of Creole} Creolists have long debated whether Creole languages have an exceptional status among the world's languages \cite{degraff2006Creole}. \newcite{mcwhorter1998} argue that Creoles are {\em simpler} than other languages, and defined by  minimal usage
of inflectional morphology, little or no use of tone encoding lexical or syntactic
contrasts, and generally semantically transparent derivation. Others have argued that Creoles cannot be be unambiguously distinguished from non-Creoles on strictly structural, synchronic grounds \cite{degraff2006Creole}. On this view Creole grammars do not form a separate typological class, but exhibit many similarities with the grammars of their parent languages, e.g., the similarities in lexical case morphology between French and Haitian Creole. We do not take sides in this debate, but observe that the exceptionalist position would explain our results that zero-shot transfer to Creole languages is particularly difficult. Exceptionalism also aligns well with the heatmaps presented in \S\ref{sec:wals}.

\paragraph{Information Bottleneck} The Information Bottleneck principle \cite{Tishby99theinformation} is an information-theoretic framework for extracting output-relevant representations of inputs, i.e., compressed, non-parametric and model-independent representations that are as informative as possible about the output. Compression is formalized by mutual information with input. A Lagrange multiplier controls the trade-off between these two quantities (informativity and compression). Being able to compute this trade-off assumes the joint input--output distribution is accessible. The trade-off is found by ignoring task-irrelevant factors and learning an invariant representation. The intuition behind the `Ancestry Bottleneck Hypothesis' ({\bf R2}) is that invariant representations are particularly useful for Creoles (see Figure~\ref{fig:front} for an illustration).  

\section{Multilingual Training}\label{sec:training}

This section sets out to evaluate the `Ancestry Transfer Hypothesis' ({\bf R1}). To this end, we evaluate multilingual language models -- trained with a BERT architecture from scratch, but of smaller size and with less data \cite{dufter-schutze-2020-identifying} -- on Creoles such as Nigerian Pidgin or Haitian Creole. 
We compare two scenarios: \textbf{1)} a scenario in which the training languages are languages that are said to be {\em parent} or {\em ancestor} languages of the Creole, such as French to Haitian, and \textbf{2)} a scenario in which {\em random}, unrelated training languages were selected. 
To compare against Creoles, we also explore these transfer scenarios for two target non-Creoles -- Spanish and Danish -- training on languages closely related to them (i.e., as typically done in cross-lingual learning).
Table~\ref{tab:lang} lists all the transfer scenarios that we investigated. 
Our experimental protocol follows \citet{dufter-schutze-2020-identifying}, and it is described in detail below.

\begin{figure}[t!]
    \centering
    \includegraphics[width=2.7in]{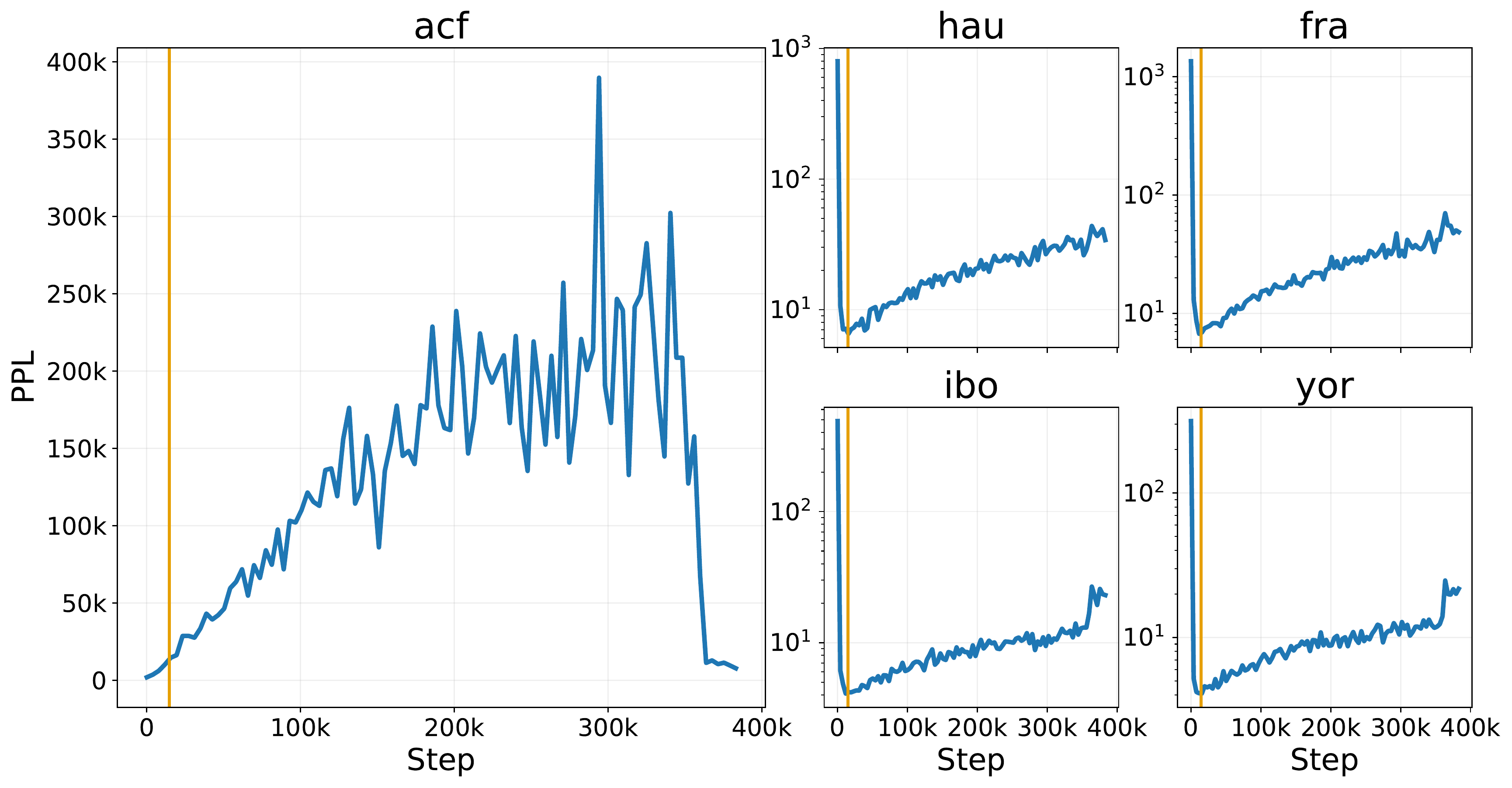}
    \includegraphics[width=2.7in]{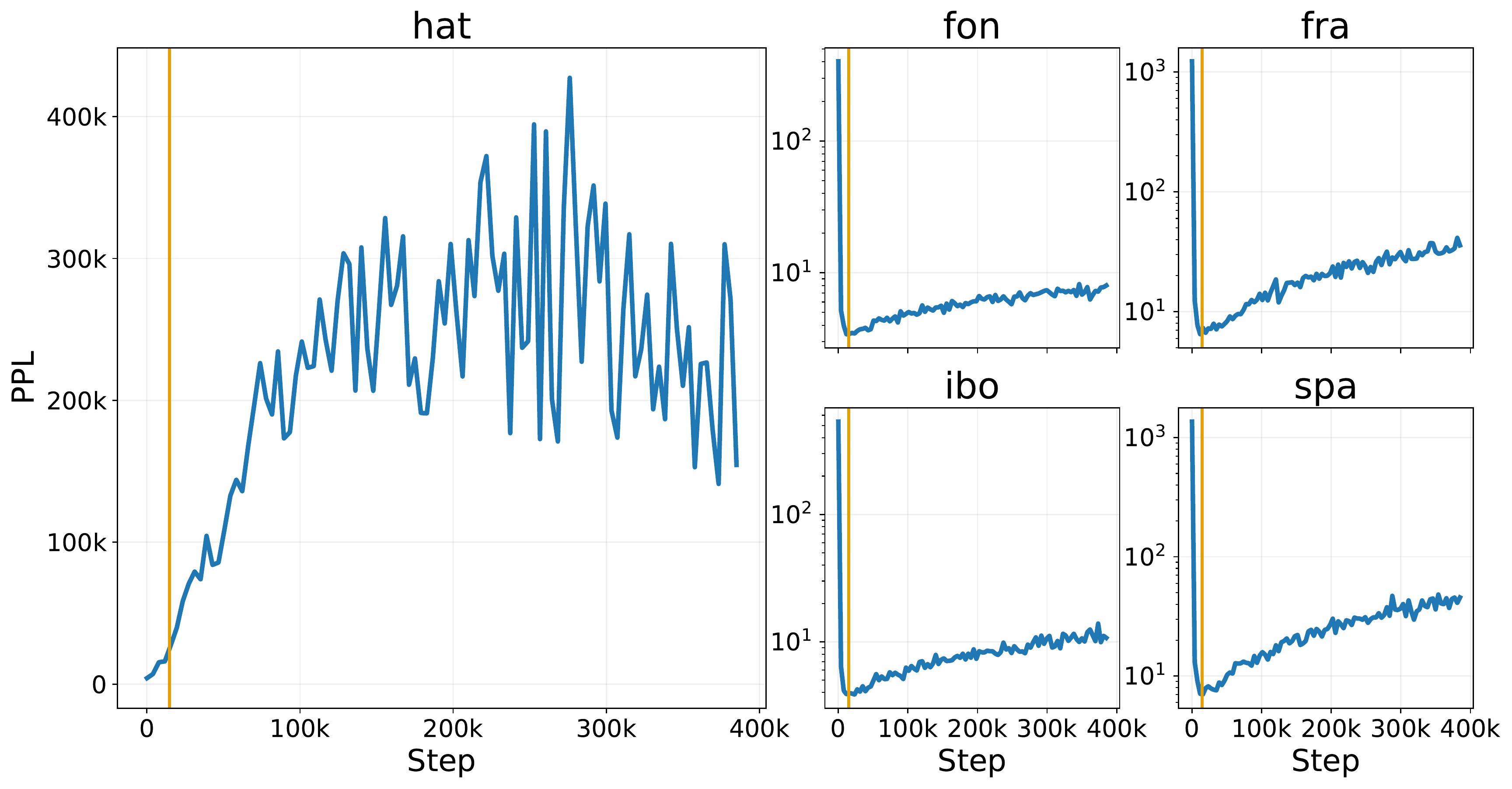}
    \includegraphics[width=2.7in]{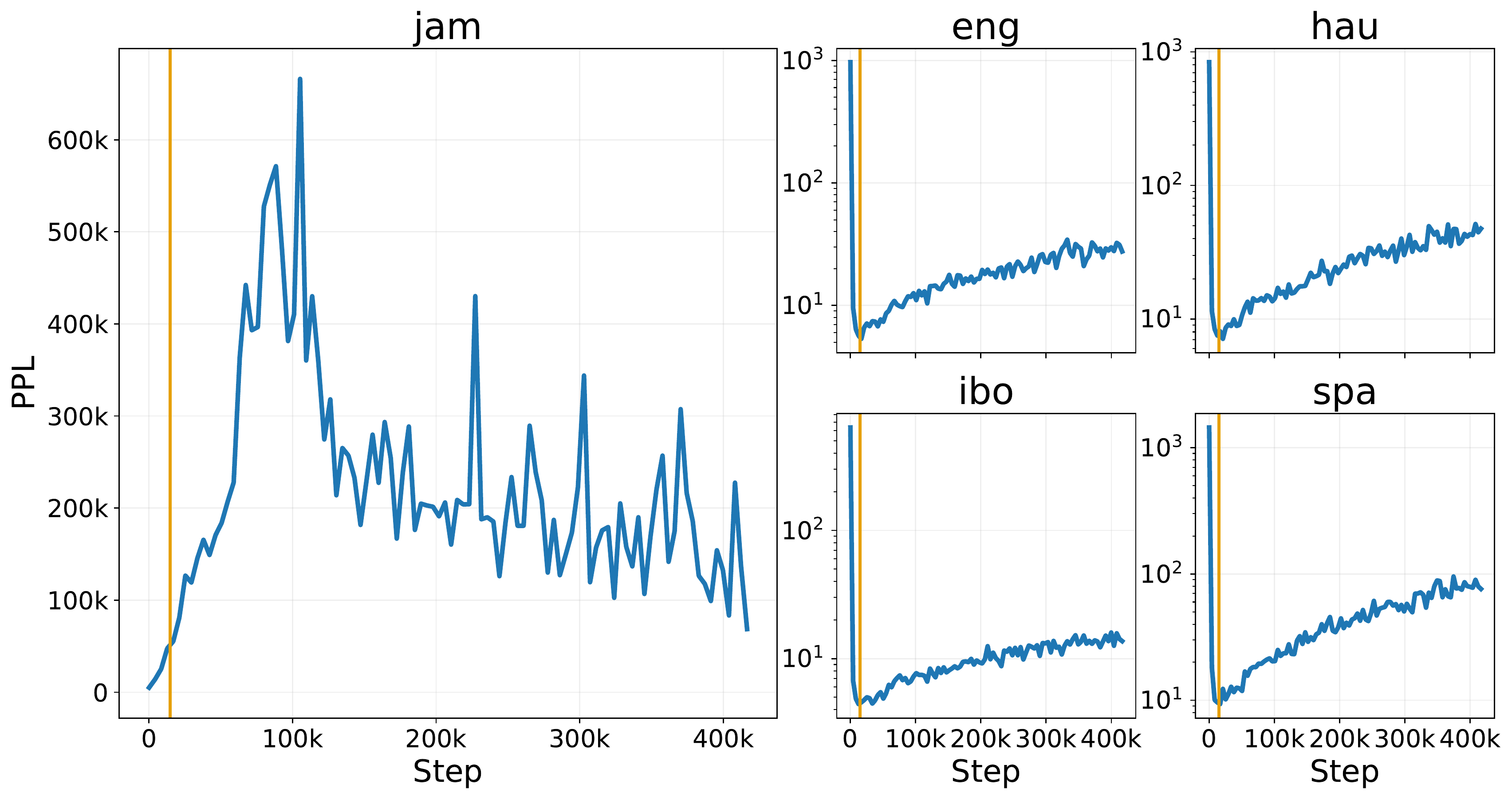}
    \includegraphics[width=2.7in]{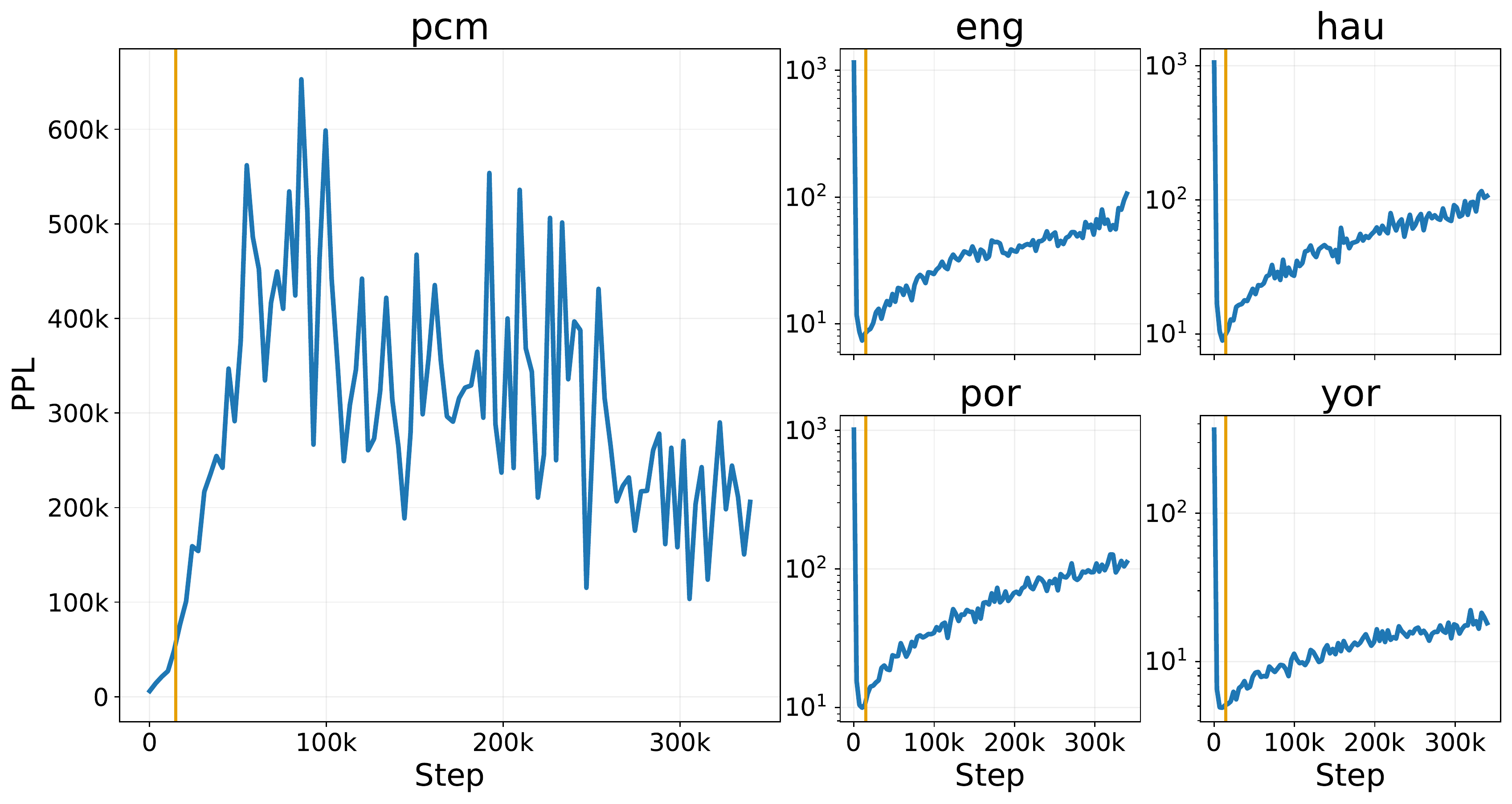}
    \vspace{-3mm}
    \caption{Four zero-shot transfer experiments for Creole languages. The left-hand side plot shows the (zero-shot) validation curve for checkpoints on Creole data; the small plots show the learning curves for the training languages. We see an initial increase in perplexity (disproving {\bf R1}). The yellow vertical line denotes 100 epochs. We also see a subsequent decrease in perplexity. 
    }
    \label{fig:learning}
    \vspace{-3mm}
\end{figure}

\begin{figure}[ht!]
    \centering
    \includegraphics[width=2.7in]{figures/pcm_parent.pdf}
    \includegraphics[width=2.7in]{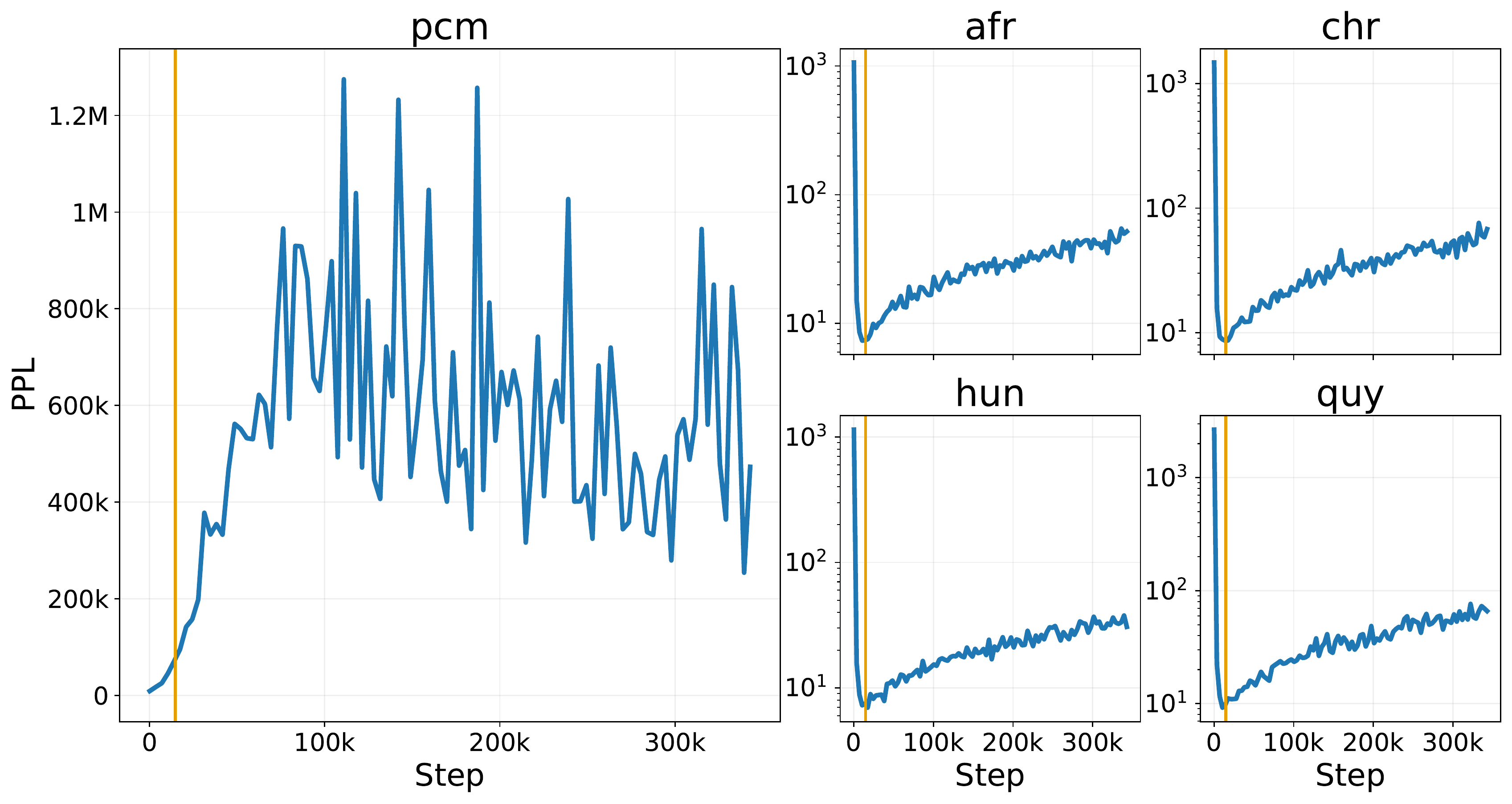}
    \vspace{-3mm}
    \caption{Learning curves for Nigerian Pidgin English when training on {\bf ancestor} languages (top) and when training on {\bf random} languages (bottom). No significant differences are observed. This disproves {\bf R2}. 
    }
    \label{fig:withrandom}
    \vspace{-3mm}
\end{figure}

We aim to learn language models for Creole languages for which large volumes of data are not readily available, and therefore explore the potential transfer from ancestor languages (the `Ancestry Transfer Hypothesis'). 
We find that standard transfer methods do not facilitate ancestry transfer. Surprisingly, different from other non-Creole languages, a very distinct two-phase pattern emerges for Creoles: As our training losses plateau, and language models begin to overfit on their source languages, perplexity on the Creoles {\em drop}. We explore if this {\em compression} phase can lead to practically useful language models (the `Ancestry Bottleneck Hypothesis'), but also falsify this. Moreover, we show that Creoles even exhibit this two-phase pattern even when training on random, unrelated languages. Thus Creoles seem to be typological outliers and we speculate whether there is a link between the two observations.

\paragraph{Experimental protocol} 
We train BERT-smaller models~\cite{dufter-etal-2020-increasing}, consisting of a single attention head (shown to be sufficient for achieving multilinguality by \citealt{K2020CrossLingualAO}). Although training smaller models means our results are not directly comparable to larger models like mBERT or XLM-R \cite{DBLP:journals/corr/abs-1911-02116}, there is evidence to support that smaller transformers can work better for smaller datasets \cite{susanto-etal-2019-sarahs}, and that the typical transformer architecture would likely be overparameterized for our small data \cite{DBLP:journals/corr/abs-2001-08361}. Thus, the BERT-smaller models appear to be the most appropriate match for our very small datasets.
The models are trained on a multilingual dataset, consisting of an equal parts of each source language, taken from the Bible Corpus \cite{mayer-cysouw-2014-creating}.
We chose Bible data to train our models as it facilitates a controlled setup with parallel data in many languages whilst including our low-resource Creoles and ancestors.
For each experiment, we learn a custom BERT tokenizer on source and target languages, with a vocabulary size of 10{,}240 word pieces~\cite{wu2016google}.\footnote{We explored different vocabulary sizes (1{,}024, 2{,}048 and 10{,}240) as well as other tokenization techniques (grapheme-to-phoneme and byte-pair encodings~\citealt{sennrich-etal-2016-neural}), which did not affect the overall findings discussed below.} 
Each model is trained for 100 epochs (see Table~\ref{tab:hyperparams}).

\begin{table}[t]
\centering
\small
\begin{tabular}{@{}lrr@{}}
\toprule
\textbf{Hyperparameter}               & \textbf{Creole}   & \textbf{Non-Creole} \\ \midrule
Vocabulary size    & 10{,}240    & 10{,}240      \\
Learning rate & 1.00E-04 & 5.00E-05   \\
Weight decay  & 1.00E-03 & 1.00E-03   \\
Dropout        & 1.00E-01 & 1.00E-01  \\
Batch size        & 256  & 256  \\
\bottomrule
\end{tabular}
\caption{The hyperparameters used for target Creole and Non-Creole experiments. Vocab size, weight decay, and dropout were the same across Creole and Non-Creole experiments, however the Non-Creoles required a smaller learning rate, in order to successfully learn. All experiments were run on a TitanRTX GPU.} \label{tab:hyperparams}
\vspace{-5mm}
\end{table}

We also follow \citet{dufter-schutze-2020-identifying}'s approach of calculating the perplexity on 15\% of randomly masked tokens ($w$), with probabilities ($p$), as $\exp(-1/n \sum_{k = 1}^{n} \log(p_{w_k})).$  We calculate perplexity on held out development data for both source and target languages. Our code is available online.\footnote{\url{https://github.com/hclent/ancestor-to-creole}}

\paragraph{Results}
In Figure~\ref{fig:learning}, by 100 epochs (indicated by a yellow vertical line), we observe two different patterns for Creoles and non-Creoles.
For target Creole languages, the models are able to learn the ancestor languages, but perplexity on the held out Creoles consistently climbs. 
On the other hand, for target non-Creoles, we observe a slight initial drop in perplexity before it starts to increase as the models overfit the source languages. 

\section{Training For Longer}\label{sec:longer}
It seems linguistically plausible that training for longer on ancestor languages to learn more invariant representations should better facilitate zero-shot transfer to Creole languages. 
This is the essence of the `Ancestry Bottleneck Hypothesis' ({\bf R2}), which we explore in this section. 

\begin{figure}[ht!]
    \centering
    \includegraphics[width=3in]{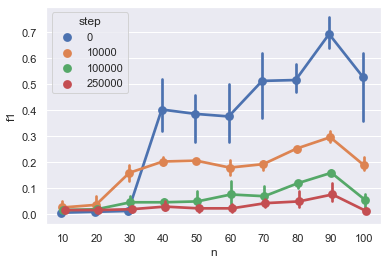}
    \vspace{-8mm}
    \caption{Results for downstream performance on Nigerian Pidgin NER, across 3 random seeds. The top row shows our model trained on ancestor of Nigerian Pidgin (pcm), while the bottom one shows results for mBERT. Step 0 in the legend refers to the pre-trained mBERT, without any further training on ancestor languages.}
    \label{fig:finetune}
    \vspace{-3mm}
\end{figure}

\paragraph{Creole compression}
We continue training our models for 5 days, for each Creole and non-Creole target language -- which typically results in 300k--500k steps of training (and thus, extremely overfit).
As the models overfit to the source languages, we observe a notable drop in perplexity for Creoles, which is true \textit{regardless} of the training data (ancestors versus random controls), as shown in Figure~\ref{fig:learning} and Figure~\ref{fig:withrandom}. 
On the other hand, these plots show that this compression does not emerge for non-Creole target languages, as their complexity steadily increases as the models overfit their training data more and more. 

\paragraph{Downstream performance}
Next, in order to determine if this compression present for Creoles can be beneficial, we used MACHAMP~\cite{vandergoot-etal-2021-machamp} to check the ability of our Nigerian Pidgin models to fine-tune for downstream NER~\cite{adelani-etal-2021-masakhaner}.
We evaluate the representations learned at different stages of pre-training by fine-tuning our checkpoints corresponding to early stage (10{,}000 steps), maximum perplexity, and post-compression (last checkpoint).
Each model is fine-tuned for 10 epochs. 
Figure~\ref{fig:finetune} shows that, across three random seeds, post-compression checkpoints consistently perform worse than pre-compression or max-complexity checkpoints. 
The results negate {\bf R2}, i.e., that the compression effect observed during training would be useful for Creoles.\footnote{We also compared the results of a pre-trained mBERT, which, unsurprisingly, outperformed all of our checkpoints (corresponding to smaller models learned from tiny data).}

\paragraph{Few-shot learning}
Finally, we assess the ability of our models to learn Creoles from few examples (n=10, ..., 100) at different training stages.
Once again, few-shot learning from post-compression checkpoints led to higher perplexity than training from maximum perplexity or early checkpoints.

\begin{figure}[t]
    \centering
    \includegraphics[width=2.1in]{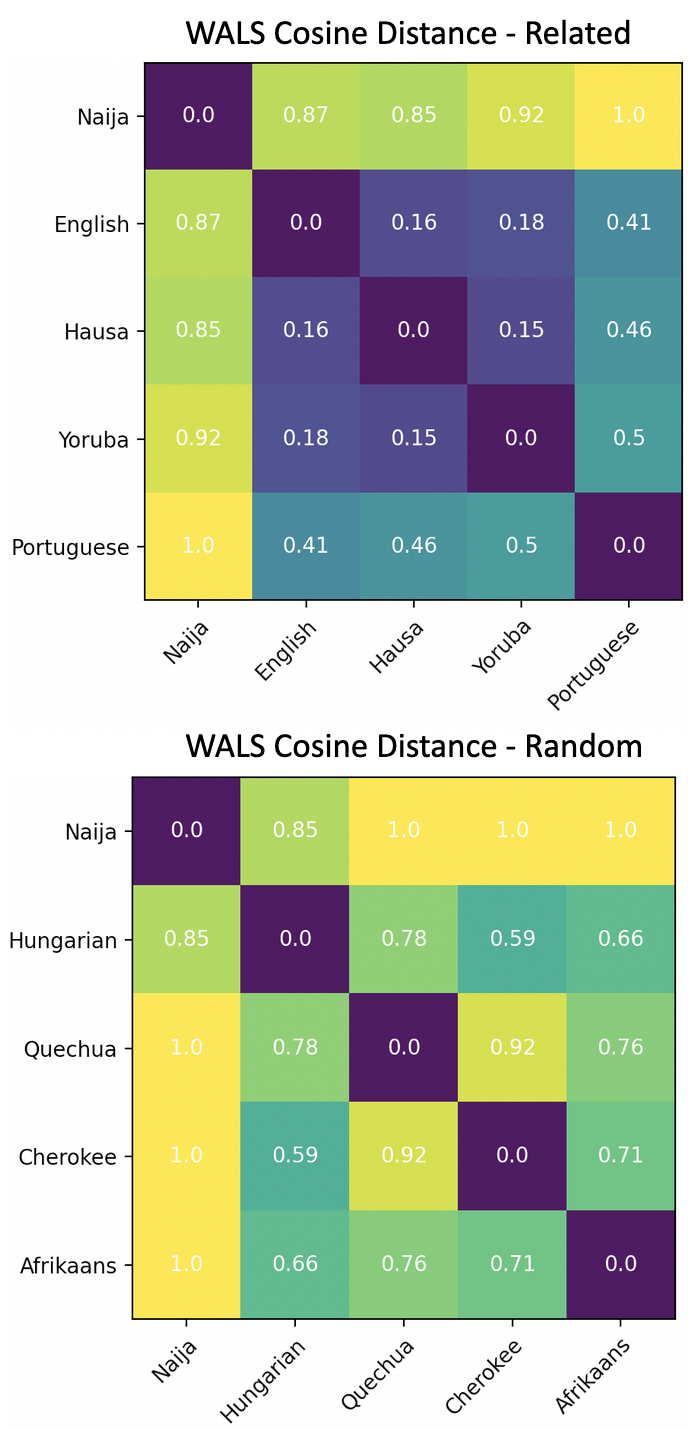}
    \caption{Heatmaps of WALS cosine distances between Nigerian Pidgin (Naija) and its parent and random training languages. We observe that Nigerian Pidgin is {\em less} related to any of these languages, than any of them internally (except Quechua and Cherokee).}
    \label{fig:heatmaps}
    \vspace{-3mm}
\end{figure}

\section{Creoles through the Lens of WALS}\label{sec:wals}

We have observed unique patterns for Creoles.
Namely, multilingual learning of the related languages did not lead to successful transfer to Creoles; and that Creoles exhibit a unique compression effect.
Here, we speculate whether there is a link between these observations, and investigate whether typological features can shed lights into our results. 
To that effect, we use The World Atlas of Language Structures (WALS)\footnote{\url{wals.info}.}, which has been used to study Creoles before~\cite{DavalMarkussen2012ExplorationsIC}.
Here, we use the cosine distance between the normalized (full) WALS feature vectors as our distance metric.\footnote{\url{https://github.com/mayhewsw/wals}.}

In Figure~\ref{fig:heatmaps}, we present an example heatmap for Nigerian Pidgin, which shows that Nigerian Pidgin is {\em less} related to ancestor and random languages than any of them internally (except Quechua and Cherokee). We found this pattern present for each of the Creoles. Thus, it would seem that Creoles' relatively large distance\footnote{We note that previous work has suggested that WALS features alone may be insufficient for typological comparison of Creoles to non-Creoles \cite{Murawaki2016StatisticalMO}.} from other languages may make cross-lingual transfer a particular challenge for learning Creoles.\footnote{We also note that cosine distance might not be meaningful here, as the normalized (full) space does not represent the feature geometry of the space that the linguists that developed the features in WALS were assuming.}

\section{Conclusion}\label{sec:conclusion}

We have presented two hypotheses ({\bf R1} and {\bf R2}) about the possibility of zero-shot transfer to Creoles, both built on the idea that Creoles share characteristics with their ancestor languages. This is not exactly equivalent to the so-called superstratist view of Creole genesis, which maintains that Creoles are essentially regional varieties of their European ancestor languages, but if the superstratist view was correct, {\bf R1} would very likely be easily validated \cite{singh-etal-2019-bert}. Our results show the opposite trend, however. Zero-shot transfer to Creole languages from their ancestor languages is hard. We do not claim that our results favor an exceptionalist position on Creoles. 
While we performed a first analysis of several segmentation approaches (i.e., BERT word piece, grapheme-to-phoneme, and byte-pair encodings) -- which did not change the training dynamics -- we believe that a rigorous comparison would be beneficial for future work in ancestor-to-Creole transfer.
We hope that continued investigation in this direction can shed more light on cross-lingual transfer, especially with regards to Creoles, and that this work has demonstrated that not all transfer between related languages is trivial.

\section{Acknowledgments}
{\small \euflag} We would like to thank the reviewers for their feedback on this manuscript. This project has received funding from the European Union’s Horizon 2020 research and innovation programme under the Marie Skłodowska-Curie grant agreement No 801199 (for Heather Lent and Emanuele Bugliarello) and the Google Research Award (for Heather Lent and Anders Søgaard). 

\bibliography{anthology,custom}
\bibliographystyle{acl_natbib}

\newpage
\clearpage
\appendix

\section{Training Setup}
\begin{table}[h]
\centering
\small
\begin{tabular}{@{}cccr@{}}
\toprule
\textbf{Type} & \textbf{Target} & \textbf{Training Langs} & \begin{tabular}[c]{@{}c@{}}\textbf{Train Size}\\ \textbf{(\#Sents)}\end{tabular} \\ \midrule
\multirow{4}{*}{Creole}     & acf & fra, hau, yor, ibo & 38,140 \\ \cmidrule(l){2-4} 
                            & hat & fra, fon, ibo, spa & 31,669 \\ \cmidrule(l){2-4} 
                            & jam & eng, hau, spa, ibo & 44,545 \\ \cmidrule(l){2-4} 
                            & pcm & eng, hau, yor, por & 35,189 \\ \midrule
\multirow{2}{*}{Non-Creole} & dan & nno, isl, swe, deu & 39,354 \\ \cmidrule(l){2-4} 
                            & spa & fra, por, ita, rom & 30,870 \\ \midrule
Control                     & -   & afr, chr, hun, quy & 37,398 \\ \bottomrule
\end{tabular}
\caption{Details of the data used for training our experiments. The same dataset was used to train "Control" experiments, for every Target language in this table. For the Train Size, the \#sents is determined by taking the parallel bible verses for each of the Training Lang(uage)s, and using a sentence splitter to obtain the training examples. All experiments had a Dev Size of 500 bible verses ($\approx500$ sentences), for all languages (Target+Training).}
\label{tab:data-stats}
\end{table}

\section{Results}

\begin{figure}[ht!]
    \centering
    \includegraphics[width=2.7in]{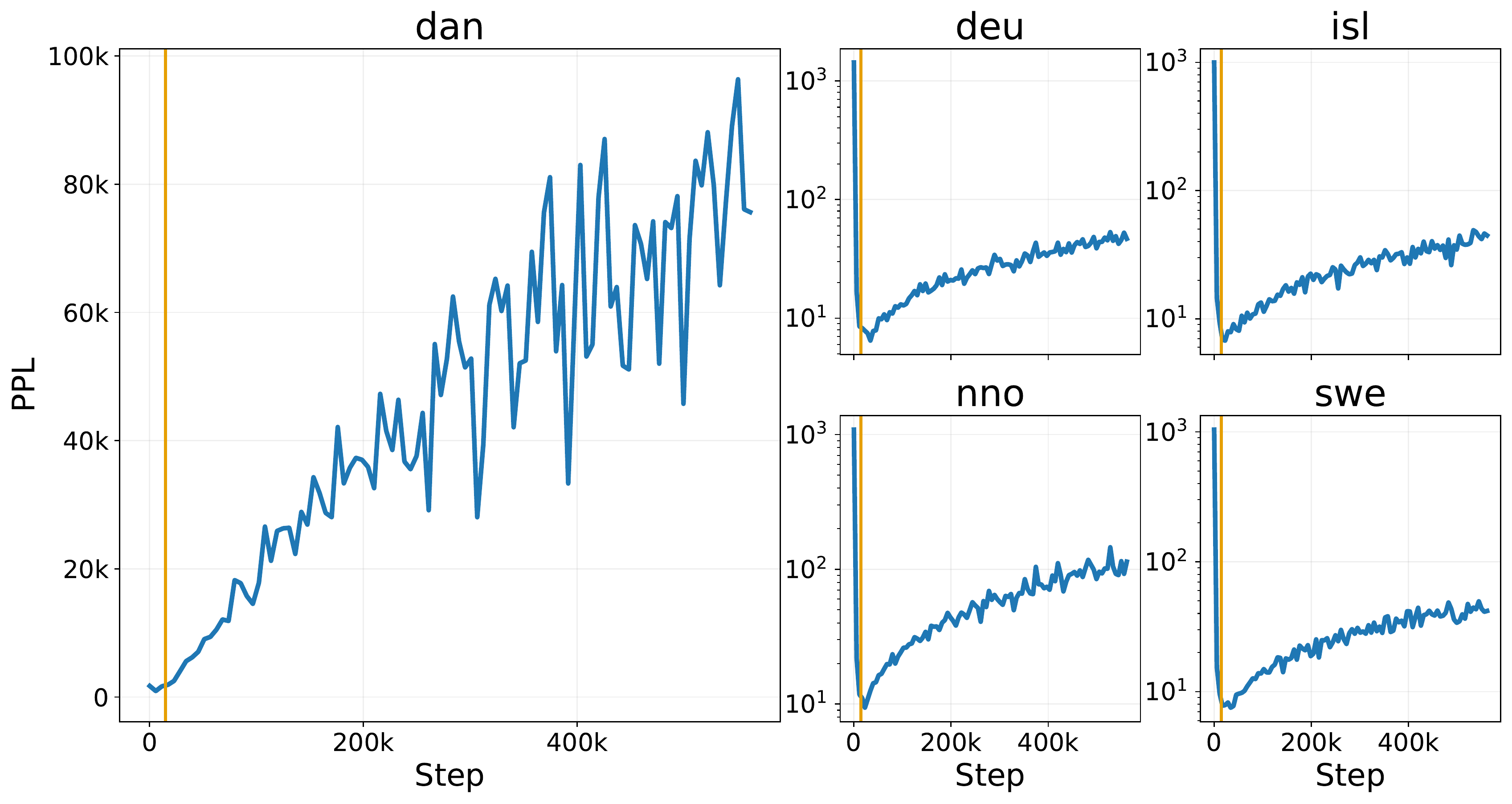}
    \includegraphics[width=2.7in]{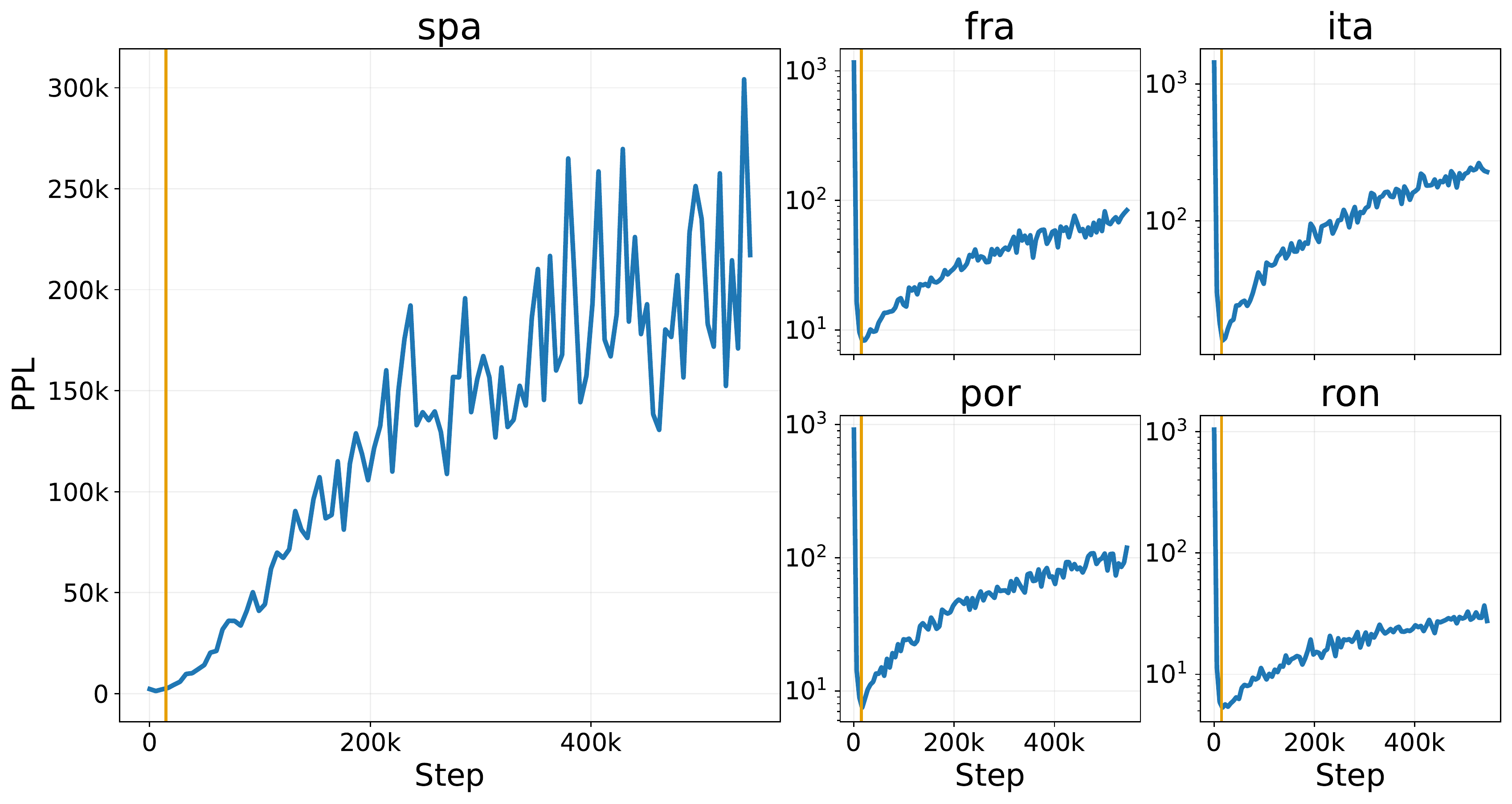}
    \caption{Full results for zero-shot transfer to non-Creole languages when training on their related languages. Before 100 epochs (shown at the yellow line), perplexity drops for the non-Creoles, as expected. As the model overfits to the training languages over time, perplexity climbs steadily.}
    \label{fig:noncreole}
    \vspace{-3mm}
\end{figure}

\begin{figure}[ht!]
    \centering
    \includegraphics[width=\columnwidth]{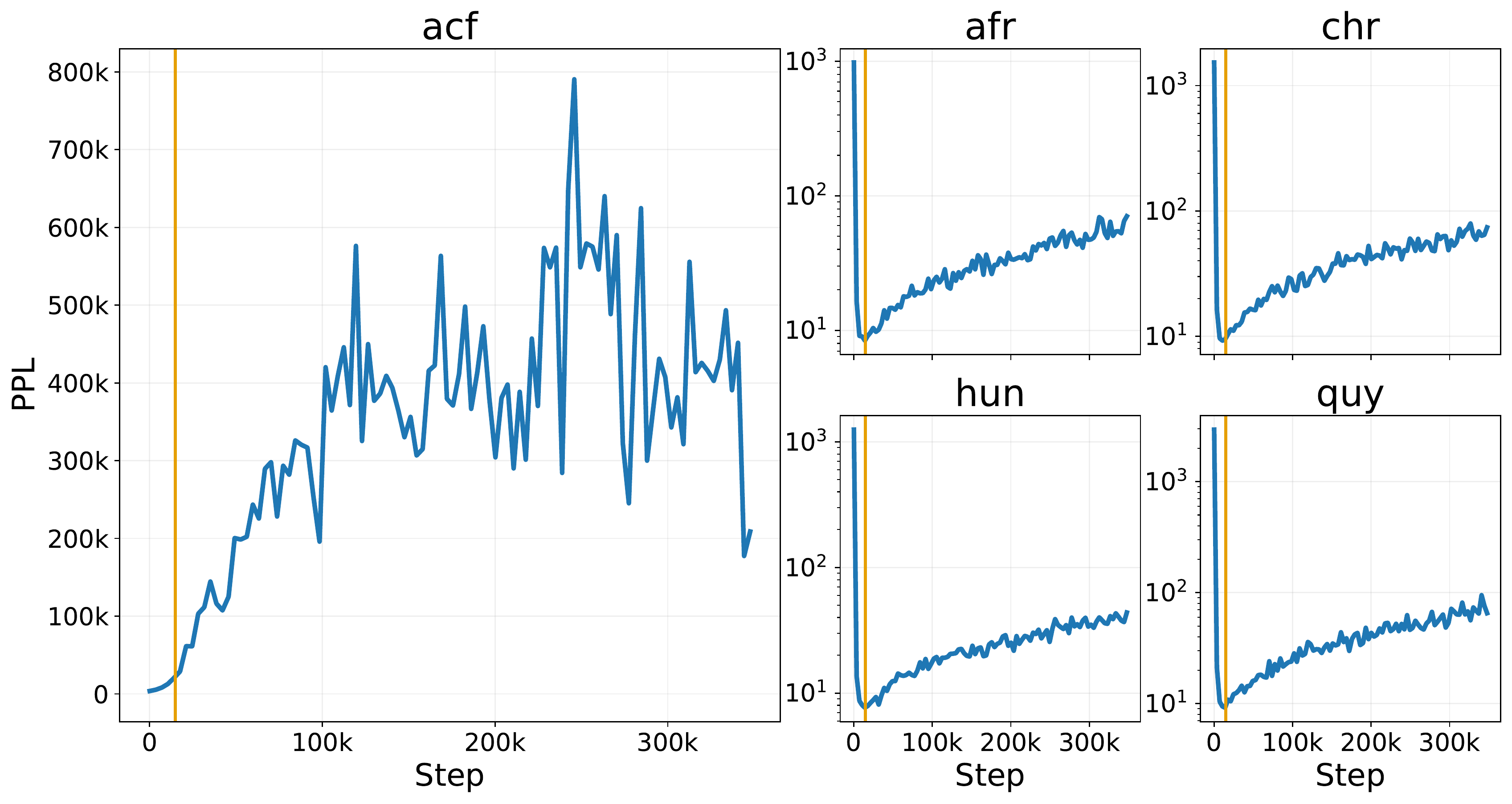}
    \includegraphics[width=\columnwidth]{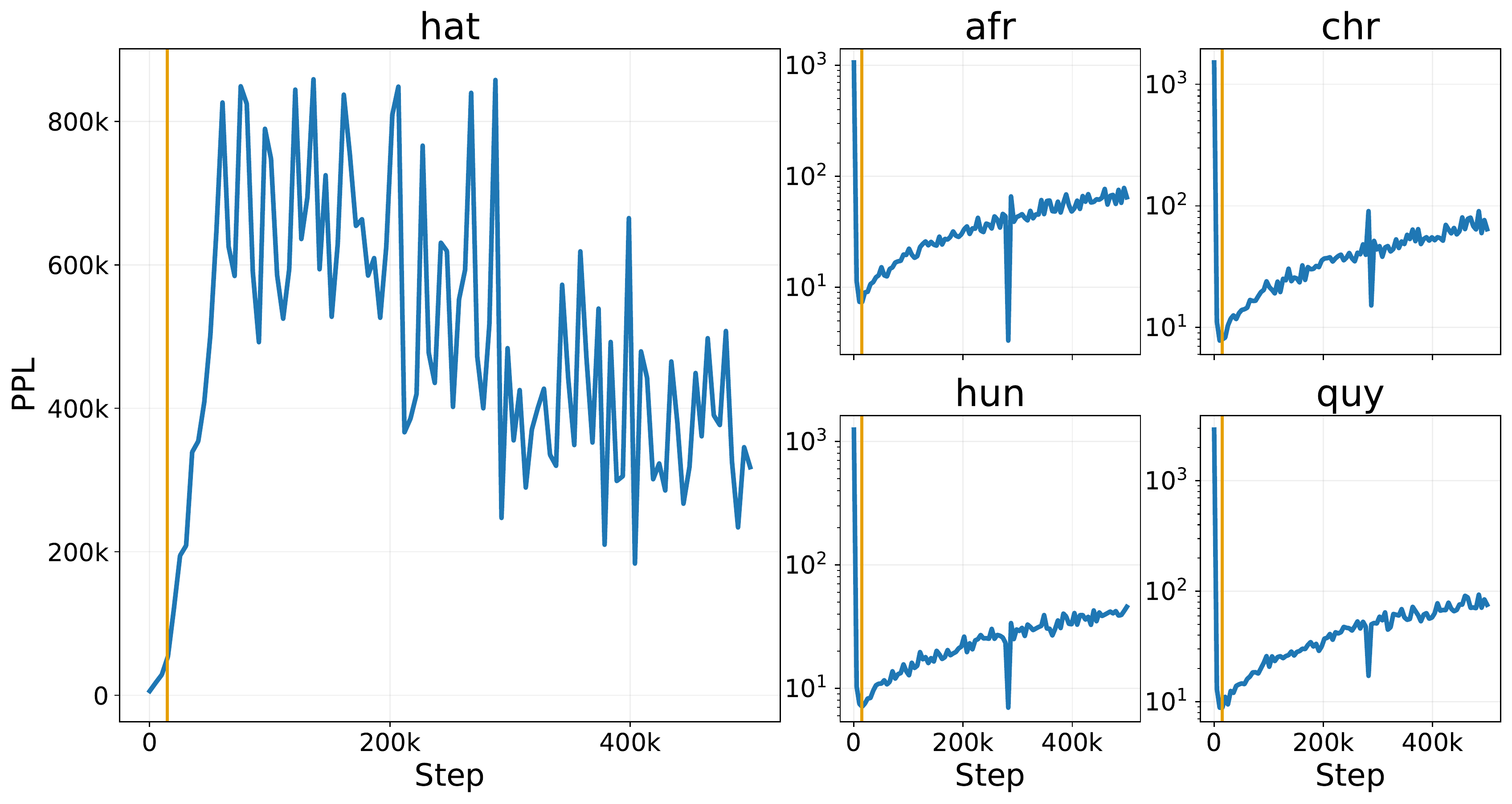}
    \includegraphics[width=\columnwidth]{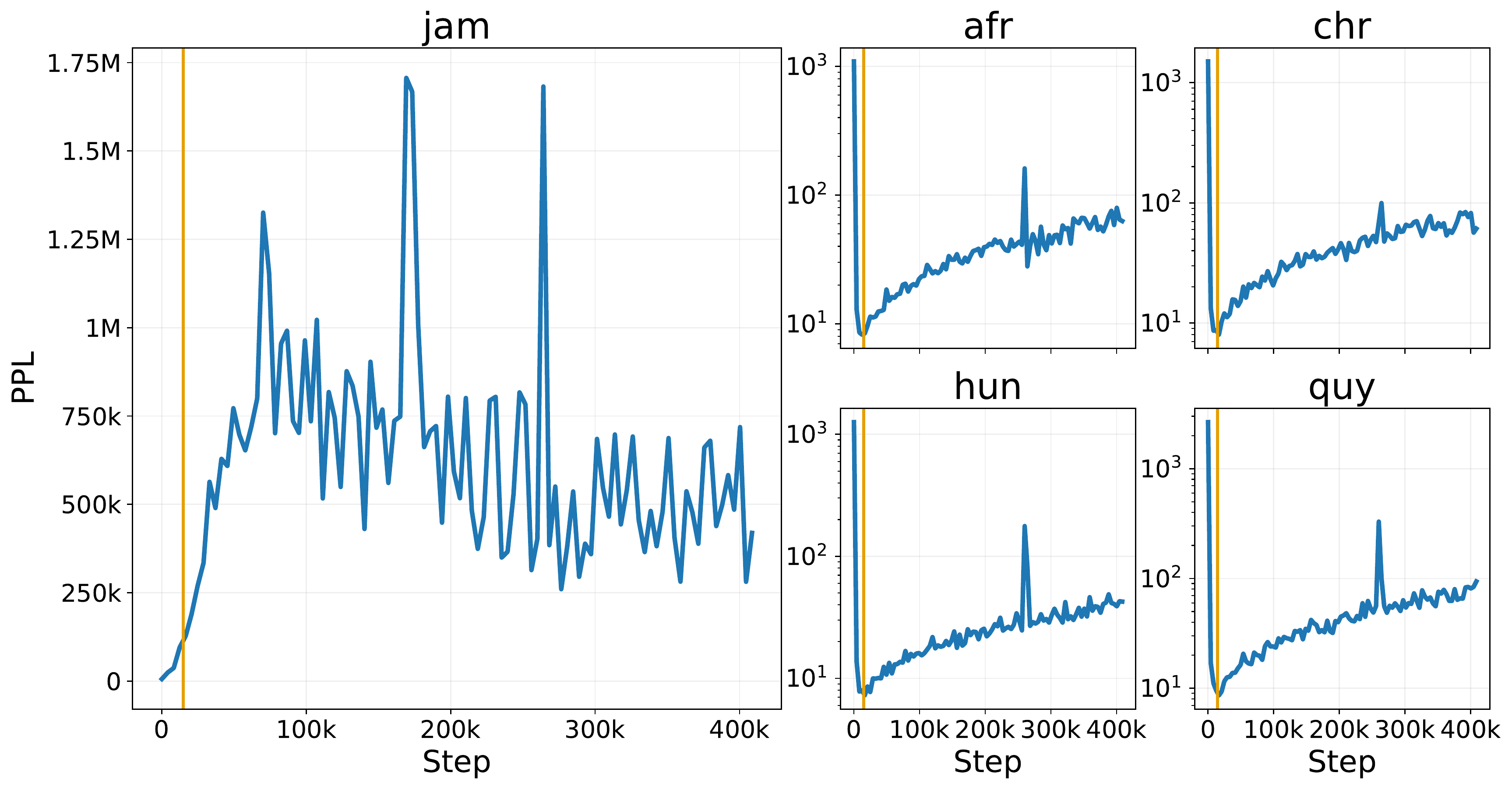}
    \includegraphics[width=\columnwidth]{figures/pcm_random.pdf}
    \caption{Full results for zero-shot transfer for Creole languages when training on random languages. The yellow line marks 100 epochs of training. Although the training languages are not related to the Creoles, we still observe the two-phase pattern, in which perplexity for Creoles drops after overfitting. }
    \label{fig:random}
    \vspace{-3mm}
\end{figure}
\vfill

\end{document}